\documentclass[preprint, 3p, 11pt, times]{elsarticle}

\usepackage{amssymb}
\usepackage{amsmath}
\usepackage{hyperref}
\usepackage{amsmath}
\usepackage{algorithm}
\usepackage{algpseudocode}
\usepackage{amsmath, amssymb, amsthm}
\usepackage{xcolor}
\usepackage{amsthm}
\usepackage{enumitem}
\usepackage{booktabs}
\usepackage{array}
\usepackage{multirow}
\usepackage{subcaption}
\newtheorem*{remark}{Remark}

\begin{document}

\begin{frontmatter}



\title{AI\textsuperscript{2}-Active Safety: \underline{AI}-enabled \underline{I}nteraction-aware Active Safety Analysis with Vehicle Dynamics}


\author[label1,label2]{Keshu Wu} 
\author[label2]{Zihao Li} 
\author[label2]{Sixu Li}
\author[label1]{Xinyue Ye}
\author[label2]{Dominique Lord}
\author[label2]{Yang Zhou\corref{cor1}} 
\affiliation[label1]{organization={Center for Geospatial Sciences, Applications, and Technology and Department of Landscape of Architecture and Urban Planning, Texas A\&M University},
            addressline={788 Ross St}, 
            city={College Station},
            postcode={77840}, 
            state={TX},
            country={United States}}
\affiliation[label2]{organization={Zachry Department of Civil and Environmental Engineering, Texas A\&M University},
            addressline={201 Dwight Look Engineering Building}, 
            city={College Station},
            postcode={77843}, 
            state={TX},
            country={United States}}

\cortext[cor1]{\raggedright Corresponding authors: Yang Zhou (\href{mailto:yangzhou295@tamu.edu}{yangzhou295@tamu.edu})}

\begin{abstract}
This paper introduces an AI-enabled, interaction-aware active safety analysis framework that accounts for groupwise vehicle interactions. Specifically, the framework employs a bicycle model—augmented with road gradient considerations—to accurately capture vehicle dynamics. In parallel, a hypergraph-based AI model is developed to predict probabilistic trajectories of ambient traffic. By integrating these two components, the framework derives vehicle intra-spacing over a 3D road surface as the solution of a stochastic ordinary differential equation, yielding high-fidelity surrogate safety measures such as time-to-collision (TTC). To demonstrate its effectiveness, the framework is analyzed using stochastic numerical methods comprising 4th-order Runge-Kutta integration and AI inference, generating probability-weighted high-fidelity TTC (HF-TTC) distributions that reflect complex multi-agent maneuvers and behavioral uncertainties. Evaluated with HF-TTC against traditional constant-velocity TTC and non-interaction-aware approaches on highway datasets, the proposed framework offers a systematic methodology for active safety analysis with enhanced potential for improving safety perception in complex traffic environments.

\end{abstract}




\end{frontmatter}




\section{Introduction}

Surrogate safety measures (SSM) serve as proactive indicators that assess potential risks in traffic scenarios before an actual collision occurs, offering a forward-looking alternative to traditional, passive active safety analyses \citep{tarko2018estimating, tarko2018surrogate, mattas2020fuzzy}.
Unlike passive approaches that rely on historical crash data and post-event assessments to gauge safety performance, surrogate measures focus on near-miss events and dynamic vehicle interactions to predict potential hazards in real time \citep{wang2021review}. This shift from reactive to predictive analysis enables active safety systems to intervene earlier, thereby enhancing collision avoidance strategies and overall traffic safety. By quantifying risk through these surrogate metrics, engineers and system designers can identify critical safety gaps and implement preemptive measures to mitigate accidents, ultimately bridging the gap between passive monitoring and active intervention \citep{morando2018studying, gettman2008surrogate}.

There are various types of SSMs, including time-based, deceleration-based, and energy-based approaches. In highway scenarios, time-based SSMs are among the most widely applied, with the Time-to-Collision (TTC) being a prominent example. Originally introduced by \citep{hayward1972near}, TTC estimates the time remaining before a collision occurs, assuming that vehicles maintain constant speed and direction. To address this limitation, \citep{ozbay2008derivation} proposed a modified TTC (MTTC) that accounts for changes in deceleration during car-following conflicts to enhance the safety assessment based on simulation. In addition, the post-encroachment time (PET) measure, introduced by \citep{allen1978analysis}, calculates the  time gap between successive vehicles using the same conflict point on the roadway. However, these SSMs are primarily designed for longitudinal conflicts (i.e., car-following) and are less effective in handling lateral movements. These limitations become even more pronounced in safety-critical scenarios such as takeovers and lane changes cases. Consequently, several studies \citep{lu2022learning, zhang2020safety, st2013automated} have extended these measures by incorporating both longitudinal and lateral dynamics, thereby offering a more comprehensive assessment of traffic safety. 

Although some approaches account for lateral movement, conventional SSMs still face limitations that undermine the accuracy of safety assessments. These models often rely on simplified kinematic assumptions that neglect the complex dynamics of acceleration, braking, and steering, which are critical for understanding real-world vehicle behavior \citep{dai2023explicitly, rahman2018longitudinal, bharat2023vehicle, li2025adaptive}. Moreover, they inadequately capture the intricate interactions among vehicles during complex maneuvers such as merging and lane changing \citep{das2022longitudinal, lu2022learning, st2013automated}.  In addition, conventional SSMs tend to overlook the uncertainties inherent in driver behavior for human-driven vehicles and path planning for connected automated vehicles \citep{wang2021review, zhang2020safety, pei2011joint}. Together, these shortcomings emphasize the need for more sophisticated models that integrate detailed vehicle dynamics, realistic interaction patterns, and behavioral uncertainties to provide a more comprehensive evaluation of highway safety. 

Recently, \citep{li2024beyond} proposed a generic SSM that is versatile across various highway geometries and capable of incorporating vehicle dynamics with varying levels of fidelity. This framework extends traditional one-dimensional SSMs to multidimensional applications, laying a solid foundation for highly accurate evaluations in dynamic traffic conditions. However, the generic SSMs only account for basic vehicle or driver behaviors that are unchanged or predefined, without carefully considering the specific dynamics of individual vehicles or drivers and the associated uncertainties \citep{bin2024surrogate, johnsson2021relative}. Considering nuanced behavior is essential because individual vehicle movements, such as subtle differences in acceleration, braking, and steering, can significantly influence crash risks \citep{laugier2011probabilistic, tavakoli2022multimodal, azadani2022novel}. These behaviors are not isolated. They interact with the movements of surrounding vehicles, creating complex dynamics that affect the overall traffic environment \citep{wu2024hypergraph, li2024disturbances, lee2024mart}. Moreover, these interactions introduce uncertainty in predicting future trajectories, as even minor variations in behavior can lead to different outcomes \citep{zhou2023interaction, zhou2023interaction, wu2023graph, li2024multi}. Therefore, integrating a detailed understanding of both individual vehicle behavior and its interactions with others into the SSM framework is critical for accurately assessing and predicting crash risks \citep{wang2021review, wang2024surrogate}.

Recent advances in graph‐based deep learning have demonstrated the power of relational models for multi‐agent trajectory forecasting. Standard graph neural networks (GNNs) represent pairwise relations via edges but remain limited when interactions involve more than two vehicles \citep{wu2023graph, zhu2024vehicle, malawade2022spatiotemporal}. Hypergraph neural networks (HGNNs) overcome this restriction by generalizing edges into ``hyperedges'', each capable of linking any number of nodes and thus directly modeling multi‐way dependencies \citep{xu2022groupnet, lu2025hyper}. In a traffic context, a hyperedge might represent all vehicles approaching a merge zone or the group of cars executing a coordinated lane change. By propagating information through these hyperedges, HGNNs capture collective behaviors and group‐level context more faithfully than pairwise GNNs \citep{wu2024hypergraph}. Although hypergraphs have begun to show promise for representing multi‐vehicle coordination in transportation research, their integration into surrogate safety analysis remains underexplored.

To address these gaps, we propose the first SSM framework that seamlessly integrates three key components: high‐fidelity vehicle dynamics, hypergraph‐based interaction modeling, and stochastic trajectory prediction. Our main contributions are:
\begin{enumerate}[nosep]
  \item We adopt an augmented bicycle model—enhanced with road gradient effects—to accurately capture turning and longitudinal behaviors while maintaining computational efficiency.
  \item We construct a hypergraph from historical trajectory data, using hyperedges to represent multi‐vehicle interaction groups, and apply a transformer‐based HGNN to infer group‐wise affinities and generate a discrete distribution of future trajectories.
  \item We fuse probabilistic ambient‐vehicle predictions with continuous host‐vehicle dynamics to compute a stochastic, interaction‐aware high-fidelity Time‐to‐Collision (HF-TTC) metric that reflects both multi‐way dependencies and behavioral uncertainties.
  \item We validate our framework on real‐world highway datasets, demonstrating clear improvements in prediction accuracy and early‐warning capability over state‐of‐the‐art TTC variants and GNN‐based extensions.
\end{enumerate}

In summary, our work advances surrogate safety analysis by unifying detailed vehicle dynamics, hypergraph reasoning, and uncertainty quantification into a cohesive framework. This integration enables more accurate and robust early‐warning assessments in complex, multi‐agent driving environments, paving the way for next‐generation active safety systems.

\section{Problem Statement}

Ensuring active safety in complex traffic environments requires a framework that is both physically grounded and capable of reasoning about multi‐agent interactions under uncertainty. In this work, we consider a \emph{host vehicle}, indexed by \(i\), traversing a three‐dimensional roadway and evaluate its potential conflict with each \emph{ambient vehicle}, indexed by \(j\). To capture realistic host‐vehicle behavior, we adopt an augmented bicycle model that incorporates steering dynamics, longitudinal acceleration, and road‐grade effects. Simultaneously, we leverage a hypergraph‐based neural network to predict the probabilistic, group‐wise trajectories of surrounding vehicles, representing each vehicle as a node and each multi‐vehicle interaction as a hyperedge. By fusing these two components, our framework quantifies surrogate safety measures—most notably a stochastic, interaction‐aware high-fidelity time‐to‐collision (HF-TTC)—that reflect both vehicle dynamics and the uncertainties inherent in driver and automated‐vehicle behaviors.

\begin{figure}[!ht]
    \centering
    \includegraphics[width=1.0\textwidth]{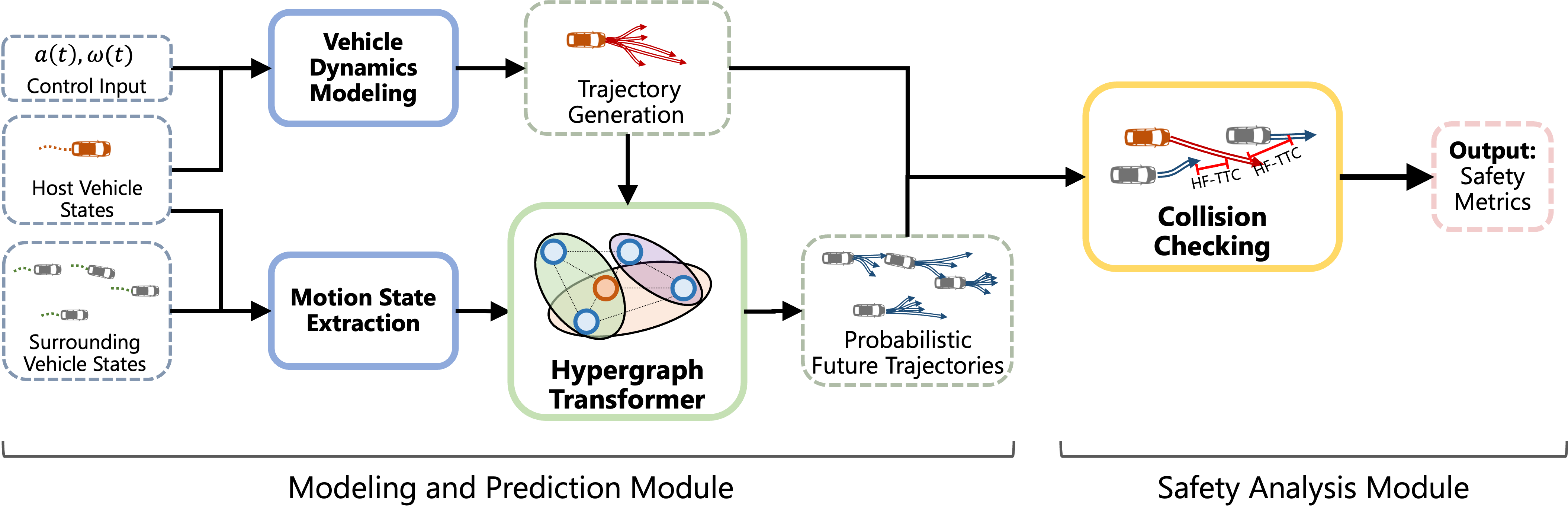}
    \caption{Framework Architecture.}
    \label{fig_framework}
\end{figure}

Figure~\ref{fig_framework} depicts the overall architecture of the proposed framework. It comprises two primary modules: a \emph{Modeling and Prediction Module} and a \emph{Safety Analysis Module}. At the core of the Modeling and Prediction Module, the host’s current state—position, speed, heading, steering angle, acceleration, and road‐gradient angle—is combined with the recent trajectory histories of nearby vehicles. A Motion State Extraction unit embeds each vehicle’s history into latent node features and computes pairwise affinities to infer a binary incidence matrix that encodes which vehicles co‐occur in interaction groups. This hypergraph structure is then processed by a multi‐layer Hypergraph Transformer, which alternately updates node and hyperedge representations to capture both individual dynamics and collective maneuvers. The transformer’s final node embeddings are decoded by multiple MLP heads to yield \(M\) candidate trajectories per vehicle, each associated with a learned probability. In parallel, the host’s future motion is generated by integrating its bicycle model via a fourth‐order Runge–Kutta scheme, under varied control‐input hypotheses, producing continuous, high‐fidelity host trajectories.

The outputs of these parallel streams feed directly into the Safety Analysis Module, where deterministic collision checks are performed between the continuous host trajectories and each discrete ambient trajectory hypothesis. By monitoring longitudinal and lateral separations against predefined safety thresholds, we extract a high-fidelity time‐to‐collision (HF-TTC) for each mode. Aggregating across the trajectory probabilities produces a stochastic HF-TTC distribution, whose inverse (HF-ITTC) cumulative distribution function offers a rigorous, interaction‐aware measure of collision risk. Together, these components deliver early‐warning safety metrics that marry realistic vehicle dynamics with probabilistic, group‐level anticipation—enabling proactive intervention far beyond what conventional, pairwise HF-TTC can achieve.



\section{Methodology}
\subsection{Vehicle Dynamics: Bicycle Model with Gradient}
\label{sec:bicycle}

To accurately represent the motion of the host vehicle in a three-dimensional environment (including road gradients), we adopt an augmented bicycle model. This model provides a balance between physical fidelity and computational efficiency. In the following, we detail the model's equations, their physical meanings, and the rationale behind each formulation.

\subsubsection{Host Vehicle Dynamics}
The host vehicle is modeled in the Cartersian coordinate system, whose position is governed by
\begin{equation}
\label{eq:dot_px}
\dot{p}_x(t) = v(t)\cos\bigl(\psi(t)\bigr), \quad \dot{p}_y(t) = v(t)\sin\bigl(\psi(t)\bigr),
\end{equation}
where:
    \(p_x(t)\) and \(p_y(t)\) denote the vehicle’s position along the \(x\)- and \(y\)-axes, respectively;
    \(v(t)\) is the instantaneous speed; and
  \(\psi(t)\) is the heading angle, determining the direction of motion.

In the bicycle model, we further consider the heading dynamics which captures the way the steering input \(\delta(t)\) induces a change in the vehicle's orientation, with the tangent function scaling the steering effect appropriately, which is described by:
\begin{equation}
\label{eq:dot_psi}
\dot{\psi}(t) = \omega(t)=\frac{v(t)}{L}\tan\bigl(\delta(t)\bigr),
\end{equation}
where:
 \(L\) is the vehicle's wheelbase (distance between the front and rear axles);
\(\delta(t)\) is the front steering angle.
    and \(\omega(t)\) is the angular speed.

For the longitudinal dynamics, the road gradient are further accounted by:
\begin{equation}
\label{eq:dot_v}
\dot{v}(t) = a(t) - g\,\sin\bigl(\alpha(t)\bigr),
\end{equation}
where, \(a(t)\) is the commanded acceleration (control input), \(g \approx 9.81\,\mathrm{m/s^2}\) is the gravitational constant; and \(\alpha(t)\) is the road gradient angle.
The term \(g\,\sin\bigl(\alpha(t)\bigr)\) accounts for the effect of the slope, ensuring that the dynamics capture both acceleration and deceleration due to the gradient.

For a concise representation, the state vector of the host vehicle is defined as:
\begin{equation}
\label{eq:x_i}
x_i(t) = 
\begin{bmatrix}
    p_x(t) \\
    p_y(t) \\
    \psi(t) \\
    v(t)
\end{bmatrix},
\end{equation}
The control input vector is defined as:
\begin{equation}
    u_i(t)=
    \begin{bmatrix}
    a(t)\\[1mm]
    \omega(t)
    \end{bmatrix},
\end{equation}
Thus, the complete state-space representation is:
\begin{equation}
\label{eq:state_space}
\frac{d}{dt}x_i(t)= f\bigl(x_i(t),u_i(t)\bigr)=
\begin{bmatrix}
v(t)\cos\bigl(\psi(t)\bigr)\\[1mm]
v(t)\sin\bigl(\psi(t)\bigr)\\[1mm]
\omega(t)\\[1mm]
a(t)-g\,\sin\bigl(\alpha(t)\bigr)
\end{bmatrix}.
\end{equation}
This formulation forms the foundation for the host vehicle trajectory planning by numerical method (e.t., Runge-Kutta 4) given below.

\subsubsection{Numerical Scheme for Host Trajectory Generation}

To generate the host vehicle's future trajectory, we integrate the state-space equations using the RK4 method. Given the initial state \(x_i(t_0)\) and a fixed time step \(\Delta t\), the state at time \(t_{n+1}=t_n+\Delta t\) is computed as:
\begin{equation}
\begin{aligned}
k_1 &= f\bigl(x_i(t_n),\, u_i(t_n)\bigr),\\[1mm]
k_2 &= f\Bigl(x_i(t_n) + \frac{\Delta t}{2} k_1,\, u_i\Bigl(t_n + \frac{\Delta t}{2}\Bigr)\Bigr),\\[1mm]
k_3 &= f\Bigl(x_i(t_n) + \frac{\Delta t}{2} k_2,\, u_i\Bigl(t_n + \frac{\Delta t}{2}\Bigr)\Bigr),\\[1mm]
k_4 &= f\Bigl(x_i(t_n) + \Delta t\, k_3,\, u_i(t_n + \Delta t)\Bigr),\\[1mm]
x_i(t_{n+1}) &= x_i(t_n) + \frac{\Delta t}{6}\Bigl( k_1 + 2k_2 + 2k_3 + k_4 \Bigr).
\end{aligned}
\end{equation}
Here, each intermediate slope \(k_\ell\) provides an estimate of the state derivative across the interval \([t_n, t_{n+1}]\). By carefully weighting these four slopes, RK4 achieves fourth‐order accuracy in \(\Delta t\), which is critical for capturing subtle nonlinearities in steering and acceleration dynamics without prohibitively small time steps.

Prior to integration, we may generate a family of plausible control sequences \(\{u_i(t)\}\) to reflect driver or planner uncertainty. Each sequence yields its own RK4 trajectory, enabling a Monte Carlo–style ensemble that propagates control variability into downstream safety metrics. In practice, we discretize controls as piecewise‐constant over each \(\Delta t\), which aligns naturally with the RK4 assumption of constant inputs within substeps.

\begin{remark}
When the control input 
\begin{equation}
u_i(t)=\begin{bmatrix} a_i(t) \\ \omega_i(t) \end{bmatrix}
\end{equation}

remains constant (i.e., \(a_i(t)=a_i\) and \(\omega_i(t)=\omega_i\) for all \(t\)), the vehicle's dynamics simplify. In this case, the commanded longitudinal acceleration \(a_i\) and the angular speed \(\omega_i\) do not vary with time. As a consequence:
\begin{itemize}
    \item The speed \(v(t)\) evolves linearly (or nearly linearly if additional terms, such as road gradient effects, are constant or negligible).
    \item The heading angle \(\psi(t)\) changes at a constant rate, with \(\dot{\psi}(t)=\omega_i\).
\end{itemize}
This constant control input assumption is commonly used in numerical integration methods, such as the Runge-Kutta 4 algorithm, where it is typically assumed that \(u_ (t)\) remains constant over each time step.
\end{remark}

By embedding this RK4 integration within our active safety pipeline, we ensure that the host’s planned trajectories faithfully reflect both high‐fidelity vehicle dynamics and control uncertainties. In the subsequent Safety Analysis Module, these continuous trajectories are directly compared against the probabilistic forecasts of surrounding vehicles to compute our stochastic Time‐to‐Collision metrics.

\subsection{Hypergraph Neural Network for Probabilistic Prediction}

\subsubsection{Input and Output}

To capture complex interactions among vehicles, we construct a hypergraph where each node corresponds to a vehicle and hyperedges represent high-order interactions among multiple vehicles. A hypergraph neural network processes this structured data to predict future trajectories for each ambient vehicle. 

\paragraph{Inputs}  Our hypergraph transformer takes as input four elements. First, the historical trajectories of all \(N\) vehicles over \(T_h\) frames, represented by
\(\mathbf{X}_h \in \mathbb{R}^{N \times T_h \times 2},\)
where each entry \(x_j(t) = [p_x^j(t), p_y^j(t)]\) encodes a vehicle’s past positions. Second, the current host‐vehicle state
\(
x_i(t_0) = \bigl[p_{x,i}(t_0),\,p_{y,i}(t_0),\,\psi_i(t_0),\,v_i(t_0)\bigr]^\top
\)
provides the starting pose and speed for trajectory generation. Third, a small ensemble of \(K\) control‐input sequences \(\{u_i^k(t)\}_{k=1}^K\) (Last-Step Constant, Average Constant, Self-Prediction) captures host‐behavior hypotheses. 

\paragraph{Outputs}  
Based on these inputs, the network produces \(M\) future trajectory hypotheses for each ambient vehicle \(j\), each with an associated probability \(p_m\). Formally, for \(m=1,\dots,M\),
\begin{equation}
x_j^{(m)}(t) =
\begin{bmatrix}
    p_{x}^{j,(m)}(t) \\
    p_{y}^{j,(m)}(t)
\end{bmatrix}, 
\quad t = t_0,\dots,t_0 + T_p,
\end{equation}
where \(T_p\) is the prediction horizon. The scalar \(p_m\) reflects the likelihood of the \(m\)-th mode and satisfies \(\sum_m p_m=1\). This discrete distribution robustly captures uncertainty in future positions.

For predicted trajectories beyond the prediction horizon, we obtain the corresponding trajectories and states by assuming a constant control input—as dictated by physical laws—with the same probability distribution. The detailed equation is given as:
\begin{equation}
\label{eq:state_space_horizon}
\frac{d}{dt}x_j^{(m)}(t)= f\bigl(x_j^{(m)}(t),u_j^{(m)}(t_0+T_p)\bigr)=
\begin{bmatrix}
v(t)\cos\bigl(\psi(t)\bigr)\\[1mm]
v(t)\sin\bigl(\psi(t)\bigr)\\[1mm]
\omega(t_0+T_p)\\[1mm]
a(t_0+T_p)-g\,\sin\bigl(\alpha(t)\bigr)
\end{bmatrix},
\quad \text{for} t>t_0+T_p.
\end{equation}
Note that Eq.~\eqref{eq:state_space_horizon} is the special case of Eq.~\eqref{eq:state_space} by assuming $a(t)$ and $\omega(t)$ remains the constant (i.e., $a(t)=a(t_0+T_p)$, $\omega(t)=\omega(t_0+T_p)$  once beyond the prediction horizon, which can be also solved by Runge-Kutta algorithm.

\subsubsection{Training versus Evaluation Protocols}

\begin{figure}[!ht]
    \centering
    \includegraphics[width=1.0\textwidth]{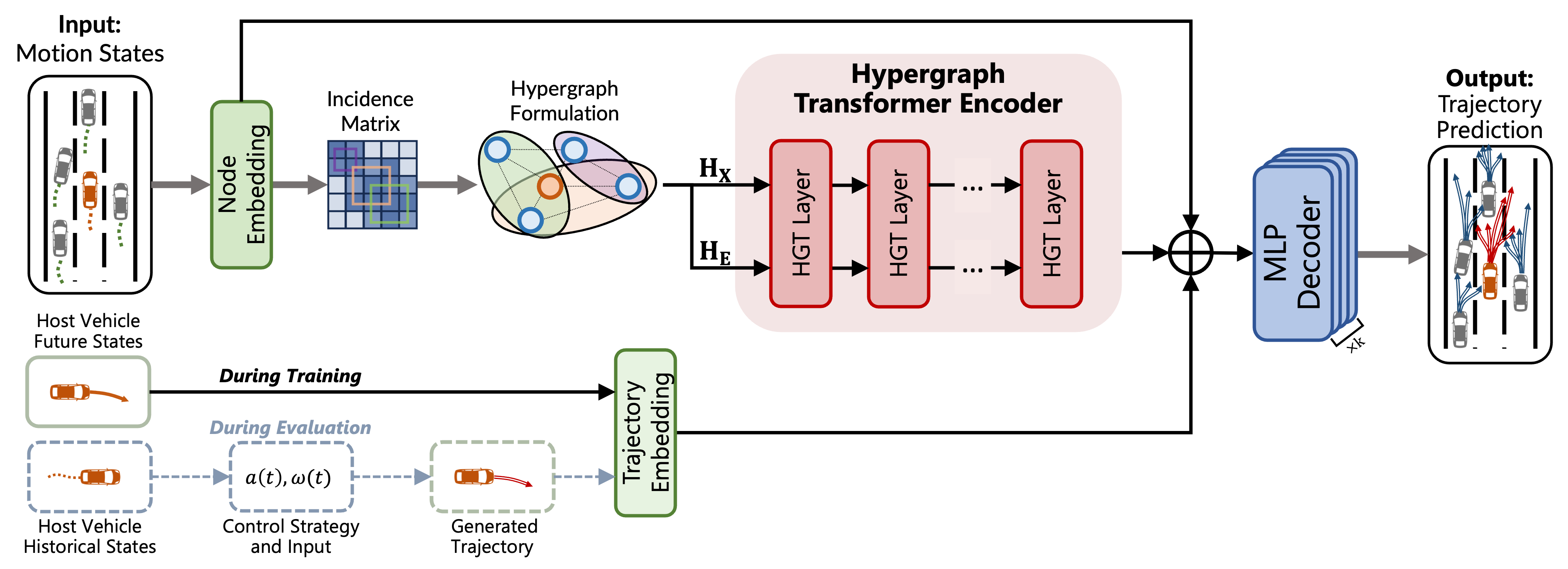}
    \caption{Hypergraph Transformer Network.}
    \label{fig_hypernn}
\end{figure}

The Hypergraph Transformer Network architecture shown in Figure~\ref{fig_hypernn} underpins our probabilistic prediction pipeline, but its host‐trajectory embedding \(Z_i\) differs between training and inference to reflect the availability of future data. In both cases, we implement this embedding via an MLP denoted \(F_{\mathrm{emb}}\).

\paragraph{Training}  
During training, we have access to the ground‐truth host trajectory \(x_i(t_0+1\!:\!t_0+T_p)\). We compute
\begin{equation}
    Z_i^{\mathrm{train}}
    = F_{\mathrm{emb}}\bigl(x_i(t_0+1\!:\!t_0+T_p)\bigr).
\end{equation}
This embedding, together with the historical encoder output and hypergraph structure, is passed through the transformer and decoder heads. The model parameters are optimized by minimizing the combined best-of-\(M\) and average MSE loss against the true future positions.

\paragraph{Evaluation}  
At inference time, the true future is unavailable. Instead, for each behavior hypothesis \(b\in\{\text{Last-Step},\text{Average},\text{Self-Prediction}\}\), we generate a candidate host trajectory \(\tilde x_i^{(b)}(t_0+1\!:\!t_0+T_p)\) via RK4. We then embed it as
\begin{equation}
    Z_i^{\mathrm{eval},b}
    = F_{\mathrm{emb}}\bigl(\tilde x_i^{(b)}(t_0+1\!:\!t_0+T_p)\bigr).
\end{equation}
This embedding replaces \(Z_i^{\mathrm{train}}\) in the transformer input, enabling the network to produce ambient‐trajectory distributions conditioned on each candidate host path—thus faithfully simulating deployment without teacher forcing.

\subsubsection{Hypergraph Representation}
A hypergraph is a natural generalization of a conventional graph, in which a hyperedge can connect more than two vertices. In our framework, vehicles are modeled as vertices and their group-wise interactions—as inferred from historical trajectory similarities—are captured as hyperedges. Formally, a hypergraph is defined as
\begin{equation}
\mathcal{H} = \{ \mathcal{V}, \mathcal{E}, A \},
\end{equation}
where $\mathcal{V}$ denotes the set of vertices, with each vertex representing an individual vehicle whose state (e.g., position, velocity) is encoded into a feature vector. The set $\mathcal{E}$ comprises hyperedges that capture multi-way interactions among vehicles, and $A$ represents the adjacency matrix that quantifies these interactions based on shared hyperedges.

\begin{figure}[!ht]
    \centering
    \includegraphics[width=0.7\textwidth]{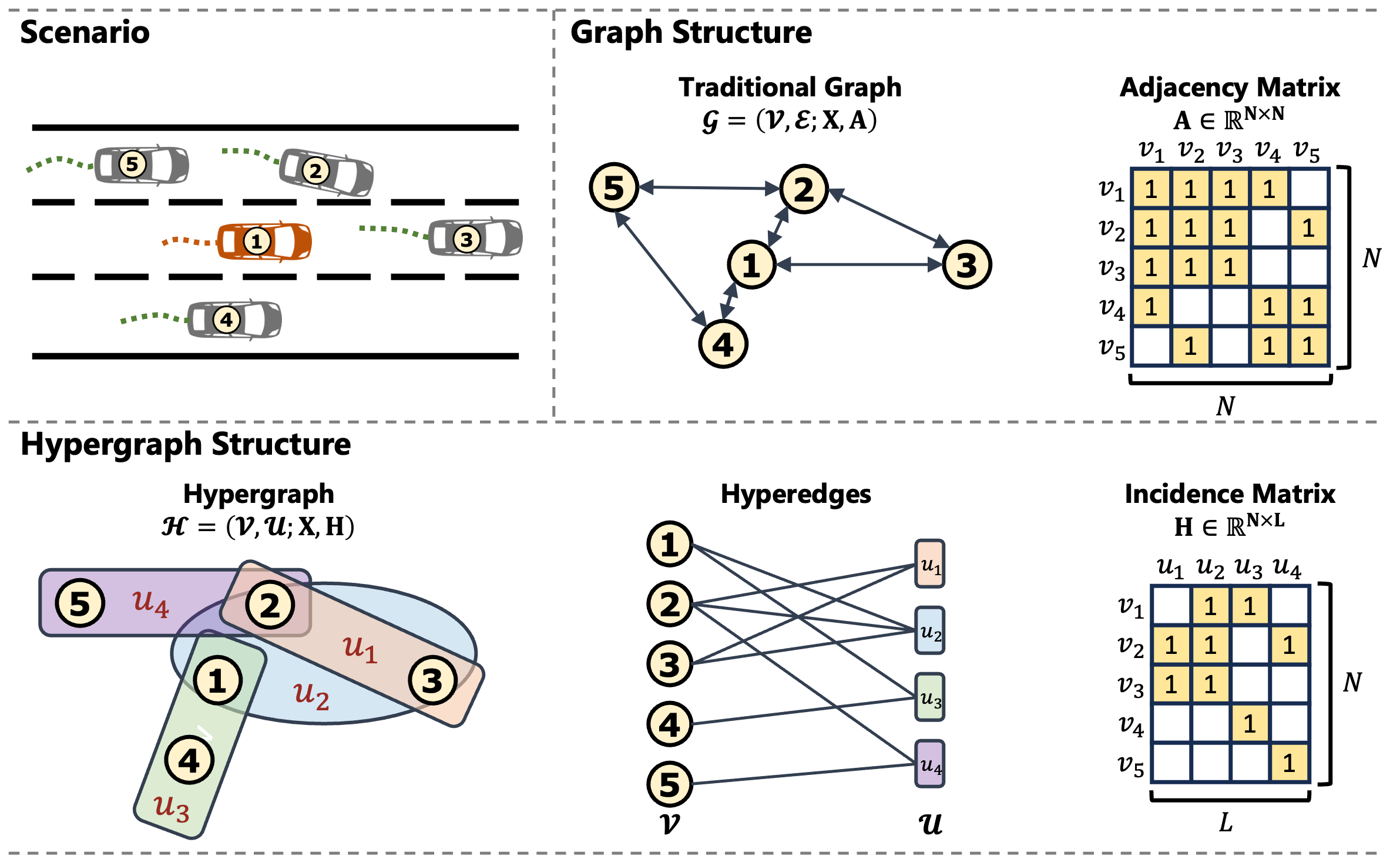}
    \caption{Graph and hypergraph.}
    \label{fig_hypergraph}
\end{figure}

One common way to represent a hypergraph mathematically is by means of its incidence matrix $H \in \{0,1\}^{|\mathcal{V}| \times |\mathcal{E}|}$. The entry
\begin{equation}
H(v,e) =
\begin{cases}
1, & \text{if } v \in e, \\
0, & \text{otherwise},
\end{cases}
\end{equation}
indicates whether vertex $v$ belongs to hyperedge $e$. For two distinct vertices $v_i$ and $v_j$, the interaction strength between them can be measured by the number of hyperedges they share, which is computed as
\begin{equation}
A_{ij} = \sum_{e \in \mathcal{E}} H(v_i,e)\,H(v_j,e).
\end{equation}
In contrast, a traditional graph restricts interactions to pairwise connections and is a special case of a hypergraph. Its adjacency matrix, denoted by $A^{(graph)}$, is defined as
\begin{equation}
A^{(graph)}_{ij} =
\begin{cases}
1, & \text{if there exists an edge connecting vertices } v_i \text{ and } v_j,\\[6pt]
0, & \text{otherwise}.
\end{cases}
\end{equation}
This fundamental difference underscores the capacity of hypergraphs to capture more complex, multi-way interactions, which is especially valuable in modeling the interactions among vehicles in dense traffic scenarios. 

Figure~\ref{fig_hypergraph} contrasts the limitations of a conventional graph representation with the expressive power of a hypergraph for modeling multi‑vehicle interactions on a highway. A standard graph connects vehicles only via pairwise edges, forcing any group maneuver—such as a three‑vehicle lane change or a platoon—to be represented as multiple independent pairwise connections. By contrast, the bottom row depicts a hypergraph in which each shaded region (hyperedge) directly groups all vehicles participating in the same collective maneuver. For instance, vehicles \(v_{1},v_{2},v_{3}\) share hyperedge \(u_{1}\) to represent their coordinated lane change, \(v_{1},v_{4}\) share \(u_{2}\) for a two‑vehicle platoon, and \(v_{2},v_{5}\) share \(u_{4}\) for a platoon. This concise grouping is encoded in the incidence matrix \(H\), where column \(u_{k}\) contains ones precisely at the rows corresponding to the vehicles in hyperedge \(u_{k}\). By capturing both pairwise and group-wise relationships in a single structure, hypergraphs enable our model to reason about complex, group‑wise interactions that traditional graphs cannot represent succinctly.

In our study, we consider the historical trajectories of both the host vehicle and the surrounding vehicles. Let 
\begin{equation}
\mathbf{X}_h \in \mathbb{R}^{N \times T_h \times d}
\end{equation}
denote the historical state matrix for all $N$ vehicles (including the host), where $T_h$ is the number of historical timesteps and $d$ is the state dimensionality (typically $d=2$ for positions $[p_x, p_y]$). For each vehicle $j$, the state at time $t$ is expressed as
\begin{equation}
x_j(t) = 
\begin{bmatrix}
p_x^j(t) \\
p_y^j(t)
\end{bmatrix}.
\end{equation}
This uniform coordinate representation facilitates consistent processing across all vehicles. A multilayer perception is used to embedded the historical trajectory of the host vehicle and each ambient vehicle embedded into a latent node feature space, yielding
\begin{equation} \label{eq:hist_embed}
n_i^{(0)} = F_{\mathrm{NI}}\bigl(x_i(1\!:\!T_h)\bigr),
\end{equation}
and 
\begin{equation}
n_j^{(0)} = F_{\mathrm{NI}}\bigl(x_j(1:T_h)\bigr),
\end{equation}
where $F_{\mathrm{NI}}(\cdot)$ is a nonlinear mapping and $d_n$ represents the dimension of the node features.

To capture the group-wise interactions among vehicles, we perform \emph{Hypergraph Topology Inference (HTI)}. First, we compute the affinity matrix $A$ by evaluating the cosine similarity between the latent node features:
\begin{equation}
A_{ij} = \frac{n_i^{(0)\top} n_j^{(0)}}{\|n_i^{(0)}\|_2 \, \|n_j^{(0)}\|_2}, \quad i,j = 1,\dots,N.
\end{equation}
This metric quantifies the similarity between the historical trajectories of vehicles, with higher values indicating more similar motion patterns. By applying a fixed threshold $\tau$, we infer the binary hypergraph topology via
\begin{equation}
G_{ij} = \mathbf{1}\{A_{ij} \ge \tau\} =
\begin{cases}
1, & \text{if } A_{ij} \ge \tau,\\[1mm]
0, & \text{otherwise}.
\end{cases}
\end{equation}
The matrix $G \in \{0,1\}^{N \times N}$ thereby encodes the group-wise connectivity among vehicles, effectively capturing which vehicles exhibit similar historical behavior and are thus likely to interact in the future. The underlying mathematical insight is that embedding trajectories into a latent space, followed by cosine similarity comparison, provides a robust measure of behavioral similarity. The threshold $\tau$ is critical in determining the granularity of the inferred interactions: lower values yield a denser hypergraph, while higher values result in a sparser connectivity pattern.

\subsubsection{Hypergraph Neural Networks}

\begin{figure}[!ht]
    \centering
    \includegraphics[width=0.5\textwidth]{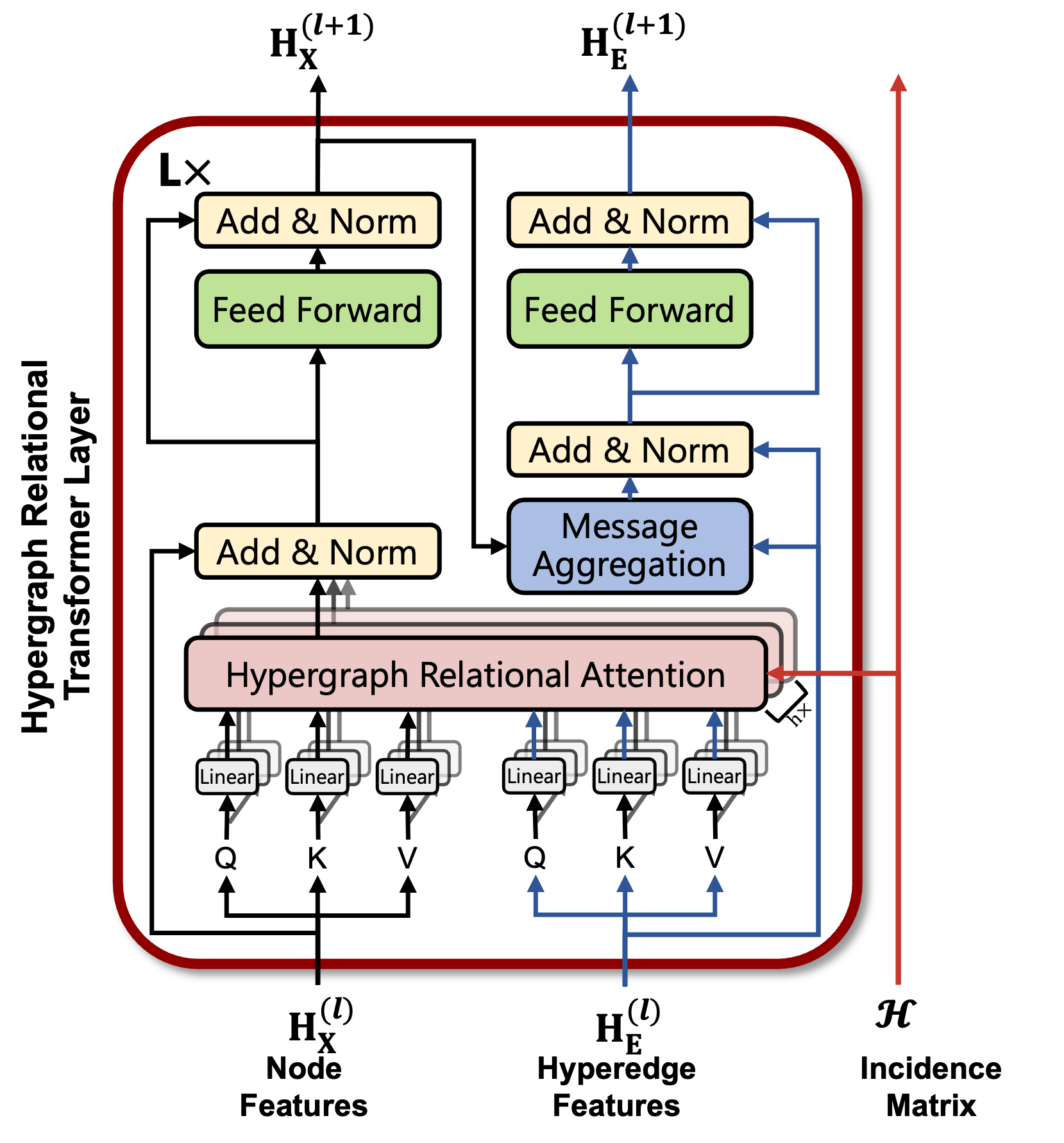}
    \caption{Hypergraph Relational Transformer.}
    \label{fig_hypergraph_relational_transformer}
\end{figure}

Once the hypergraph is established, we employ a hypergraph neural network based on a \emph{Hypergraph Transformer}, as shown in Figure \ref{fig_hypergraph_relational_transformer},  (HGT) to extract interaction-aware features for probabilistic trajectory prediction. Built based on \citep{lee2024mart}, the HGT iteratively refines both node features $n_j^{(l)}$ and hyperedge features $h_j^{(l)}$ through a series of transformer layers \citep{vaswani2017attention}. Each layer leverages the inferred topology $G$ to integrate both individual and group-level information.

In each transformer layer $l$, the update mechanism begins with an augmented self-attention module. For each node $i$, the query vector is defined as
\begin{equation}
q_i = n_i^{(l)}W^Q_n + \frac{1}{|H_i|}\sum_{j\in H_i} h_j^{(l)}W^Q_h,
\end{equation}
where $W^Q_n \in \mathbb{R}^{d_n\times d_n}$ and $W^Q_h \in \mathbb{R}^{d_e\times d_n}$ are learnable projection matrices, and $H_i = \{ j \mid G_{ij}=1 \}$ denotes the set of vehicles connected to node $i$. Likewise, the key and value vectors for node $j$ are computed as
\begin{equation}
k_j = n_j^{(l)}W^K_n + \frac{1}{|H_j|}\sum_{i\in H_j} h_i^{(l)}W^K_h,
\end{equation}
\begin{equation}
v_j = n_j^{(l)}W^V_n + \frac{1}{|H_j|}\sum_{i\in H_j} h_i^{(l)}W^V_h.
\end{equation}
The self-attention weight between nodes $i$ and $j$ is calculated as
\begin{equation}
\alpha_{ij} = \frac{\exp\Bigl(\frac{q_i \cdot k_j}{\sqrt{d_n}}\Bigr)}{\sum_{j'=1}^{N}\exp\Bigl(\frac{q_i \cdot k_{j'}}{\sqrt{d_n}}\Bigr)},
\end{equation}
which quantifies the influence of node $j$ on node $i$ by considering both its individual dynamics and the aggregated group context. The aggregated representation for node $i$ is then given by
\begin{equation}
\tilde{n}_i = \sum_{j=1}^{N}\alpha_{ij}\, v_j,
\end{equation}
and the updated node feature is obtained via a residual connection followed by layer normalization and a feed-forward network:
\begin{equation}
n_i^{(l+1)} = \operatorname{LayerNorm}\Bigl(n_i^{(l)} + \operatorname{FFN}\bigl(\tilde{n}_i\bigr)\Bigr).
\end{equation}

Simultaneously, the hyperedge features are updated to reflect the evolving group context. For a hyperedge associated with node $i$, a message vector is constructed by concatenating the current hyperedge feature with the aggregated updated node features from the group:
\begin{equation}
m_i^{(l)} = \operatorname{ReLU}\Biggl(\Bigl[\, h_i^{(l)}; \, \frac{1}{|N_i|}\sum_{j\in N_i} n_j^{(l+1)} \Bigr]W_m\Biggr),
\end{equation}
where $N_i = \{ j \mid G_{ij}=1 \}$ and $W_m$ is a learnable weight matrix. The hyperedge is then updated via a transformer-style function:
\begin{equation}
h_i^{(l+1)} = F_h\bigl(m_i^{(l)}, h_i^{(l)}\bigr),
\end{equation}
which typically includes additional residual connections and normalization.

After \(L\) layers of the Hypergraph Transformer, each node embedding \(n_i^{(L)}\) integrates both the individual vehicle’s dynamics and the group‐wise context inferred from historical trajectories. To produce probabilistic forecasts, these embeddings are combined with a trajectory embedding \(Z_i\) and the initial history embedding \(n_i^{(0)}\) defined in Eq.~\eqref{eq:hist_embed}, and then passed through multiple MLP decoder heads.

In our framework, the host‐trajectory embedding \(Z_i\) is instantiated according to the operational phase:
\begin{equation}
    Z_i =
    \begin{cases}
    Z_i^{\mathrm{train}}, & \text{during training stage},\\
    Z_i^{\mathrm{eval},b}, & \text{during evaluation stage}.
    \end{cases}
\end{equation}
We then assemble the decoder input by concatenating this embedding with both the final transformer output and the initial history encoding:
\begin{equation}
    \mathbf{h}_i^{\mathrm{dec}} = 
    \begin{bmatrix}
    n_i^{(L)} \\[2pt]
    n_i^{(0)} \\[2pt]
    Z_i
    \end{bmatrix}.
\end{equation}

Each of the \(M\) decoder heads—denoted \(F_{\mathrm{dec}}^{(m)}\)—maps \(\mathrm{Input}_i\) to both a predicted trajectory and a confidence logit:
\begin{equation}
\label{eq:decoder_concat}
\bigl[\hat{Y}_i^{(m)},\,z_i^{(m)}\bigr]
= F_{\mathrm{dec}}^{(m)}\bigl(\mathbf{h}_i^{\mathrm{dec}}\bigr),
\quad m=1,\dots,M,
\end{equation}
with \(\hat{Y}_i^{(m)}\in\mathbb{R}^{T_p\times2}\) the \(m\)-th trajectory hypothesis. We convert logits \(z_i^{(m)}\) into probabilities via softmax,
\begin{equation}
p_i^{(m)} = \frac{\exp\bigl(z_i^{(m)}\bigr)}{\sum_{m'=1}^M \exp\bigl(z_i^{(m')}\bigr)},
\end{equation}
so that \(\sum_{m}p_i^{(m)}=1\). Here, $p_i^{(m)} \in [0,1]$ represents the probability that vehicle $i$ will follow the $m$-th predicted trajectory. The confidence logit $z_i^{(m)}$ provides a quantitative measure of the reliability of each prediction. A higher logit value leads to a larger probability, indicating stronger confidence in that particular future trajectory. By concatenating the final and historical embeddings with the host‐trajectory embedding, and by employing multiple decoder heads, the model captures both high‐fidelity dynamics and interaction uncertainty, yielding a discrete distribution over plausible futures instead of a single deterministic path.

\subsection{Stochastic High-fidelity Time-to-Collision (HF-TTC) Computation}
Building on the probabilistic trajectory predictions, we next perform a stochastic safety analysis by computing a high‑fidelity TTC (HF-TTC) that accounts separately for longitudinal and lateral separations. For each host–ambient vehicle pair \((i,j)\) and for each predicted mode \(m\) with probability \(p_m\), we denote the host’s state as 
\begin{equation}
    x_i(t) = \bigl[p_{x,i}(t),\,p_{y,i}(t)\bigr]^\top,
\end{equation}
and the \(m\)-th ambient trajectory as 
\begin{equation}
    x_j^{(m)}(t) = \bigl[p_{x,j}^{(m)}(t),\,p_{y,j}^{(m)}(t)\bigr]^\top,
\end{equation}
for \(t\in[t_0,\,t_0+T_p]\). We then compute the longitudinal and lateral separations,
\begin{equation}
    \Delta x_{ij}^{(m)}(t) = \bigl|p_{x,i}(t) - p_{x,j}^{(m)}(t)\bigr|,\quad
    \Delta y_{ij}^{(m)}(t) = \bigl|p_{y,i}(t) - p_{y,j}^{(m)}(t)\bigr|.
\end{equation}
A collision is assumed when \(\Delta x_{ij}^{(m)}(t)\le r_x\) and \(\Delta y_{ij}^{(m)}(t)\le r_y\), where \(r_x\) and \(r_y\) are the longitudinal and lateral safety thresholds, respectively. The deterministic HF-TTC for mode \(m\) is then
\begin{equation}
\mathrm{HF\text{-}TTC}_{ij}^{(m)}(t_0) 
= \min\Bigl\{\,t \ge t_0 \,\Bigm|\,\Delta x_{ij}^{(m)}(t)\le r_x \;\land\;\Delta y_{ij}^{(m)}(t)\le r_y\Bigr\}.
\end{equation}
The corresponding inverse TTC (HF-ITTC) is
\begin{equation}
\mathrm{HF\text{-}ITTC}_{ij}^{(m)} = \frac{1}{\mathrm{HF\text{-}ITTC}_{ij}^{(m)}(t_0)}.
\end{equation}

\begin{figure}[!ht]
    \centering
    \includegraphics[width=0.92\textwidth]{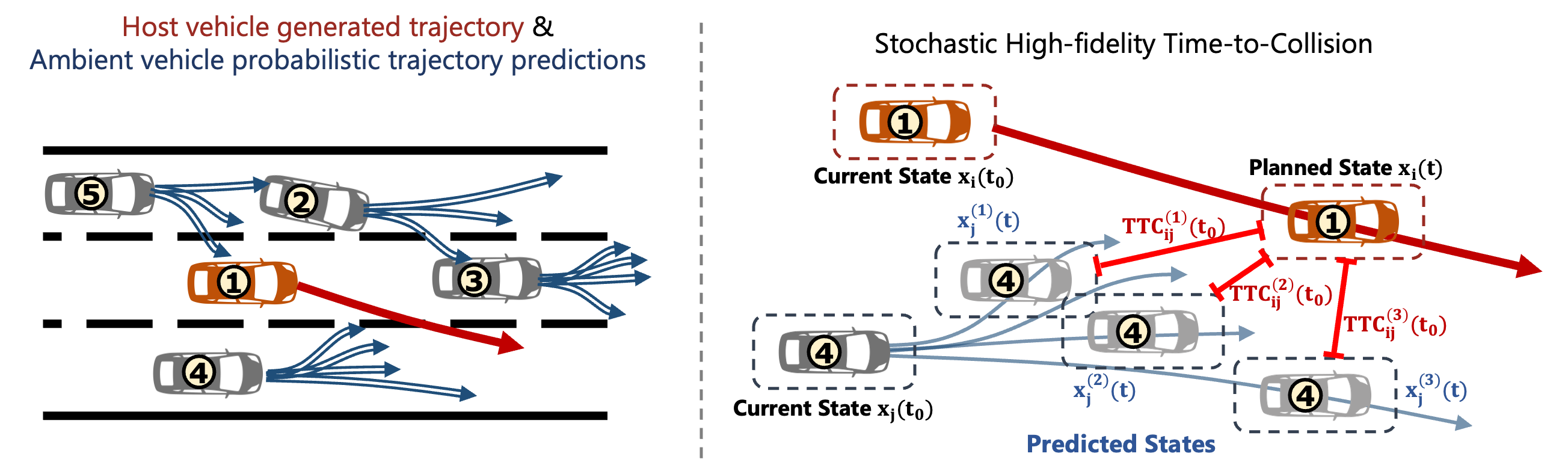}
    \caption{Stochastic High-fidelity Time-to-Collision.}
    \label{fig_ttc}
\end{figure}

As illustrated in Figure~\ref{fig_ttc}, the {\it left} panel overlays the host’s RK4‐generated trajectory with the ensemble of ambient‐vehicle forecasts.  The {\it right} panel zooms in on a single host–ambient pair: three dashed boxes show distinct ambient‐trajectory modes \(x_j^{(1)}(t)\), \(x_j^{(2)}(t)\), \(x_j^{(3)}(t)\), and the red braces mark the times at which each mode first breaches both longitudinal and lateral thresholds, thereby defining \(\mathrm{HF\text{-}TTC}_{ij}^{(m)}\).

Considering the probabilistic nature of the predictions, the overall HF-ITTC is a discrete random variable with probability mass function
\begin{equation}
f_{\mathrm{HF-ITTC}}(t) = \sum_{m=1}^{M} p_m\, \delta\Bigl(t - \mathrm{HF-ITTC}_{ij}^{(m)}\Bigr),
\end{equation}
where \(\delta(\cdot)\) is the Dirac delta function. Its cumulative distribution function (CDF) is given by
\begin{equation}
F_{\mathrm{HF-ITTC}}(t) = \sum_{m=1}^{M} p_m\, H\Bigl(t - \mathrm{HF-ITTC}_{ij}^{(m)}\Bigr),
\end{equation}
with \(H(\cdot)\) being the Heaviside step function.
By that, this approach provides a comprehensive view of the surrogate safety measures by accounting for group-wise interactions, motion uncertainty, and high-fidelity dynamics.

\section{Experiments and Analysis}

We designed a suite of four experiments to rigorously validate the core innovations of our interaction‑aware, dynamics‑driven active safety framework. These experiments respectively quantify end‑to‑end prediction accuracy, examine the effect of host vehicle's behavior models on risk prediction, assess real‑time performance under adaptive replanning, and characterize sensitivity to key modeling parameters.

\subsection{Experiments Settings}
\subsubsection{Data Preparations}
Our study employs two publicly available highway trajectory datasets for training and validation: the Next Generation Simulation (NGSIM) dataset \citep{ngsimdatasetus101,ngsimdatasetus80} and the HighD dataset \citep{krajewski2018highd}. The NGSIM data were collected via overhead cameras at 10 Hz on the eastbound I-80 corridor in the San Francisco Bay Area and the southbound US 101 in Los Angeles, providing detailed vehicle position and velocity records. The HighD dataset, captured by drone footage at 25 Hz between 2017 and 2018 near Cologne, Germany, spans approximately 420 m of bidirectional roadway and comprises about 110 000 vehicle trajectories (cars and trucks) covering a total distance of 45 000 km.

Following preprocessing, the NGSIM set contains 662 000 rows corresponding to 1 380 unique trajectories, while the HighD set includes 1.09 million rows representing 3 913 trajectories. We split each dataset into a 70\% training partition and a 30\% testing partition. For temporal context, we use a historical window of \(T = 30\) frames and forecast \(F = 50\) future frames for each trajectory.

\subsubsection{Training and Evaluation Metrics}

\paragraph{Loss Functions}  
For probabilistic, multi‑modal trajectory prediction, we employ a two‑term loss that balances accuracy of the best hypothesis with coverage of all modes. Let \(Y_i(t)\in\mathbb{R}^2\) be the ground‑truth position of vehicle \(i\) at future time \(t\), and let \(\hat{Y}_i^{(m)}(t)\) for \(m=1,\dots,M\) denote the \(M\) predicted trajectories with associated probabilities \(p_i^{(m)}\). The training loss for each vehicle \(i\) is
\begin{equation}
    \mathcal{L}_i
    =\underbrace{\min_{m\in\{1,\dots,M\}}\sum_{t=1}^{T_f}\bigl\|Y_i(t) - \hat{Y}_i^{(m)}(t)\bigr\|_2^2}_{\text{best‑of‑}M\;\text{MSE}}
    \;+\;\lambda
    \underbrace{\frac{1}{M}\sum_{m=1}^{M}\sum_{t=1}^{T_f}\bigl\|Y_i(t) - \hat{Y}_i^{(m)}(t)\bigr\|_2^2}_{\text{average MSE over }M\;\text{modes}},
\end{equation}
where \(\lambda=1\). The first term ensures that at least one mode closely matches the ground truth, while the second term encourages all modes to remain plausible. Deterministic baselines optimize the single‑mode MSE \(\sum_{t=1}^{T_f}\|Y_i(t) - \hat{Y}_i(t)\|_2^2\).

\paragraph{Evaluation Metrics}  
To rigorously assess the quality of our multi‐modal trajectory forecasts, we evaluate four complementary error measures on the held‐out test set. Each metric quantifies a different aspect of prediction accuracy—ranging from per‐step fidelity to end‐point precision and overall error distribution—thereby providing a holistic view of model performance.

\begin{itemize}[leftmargin=*]
  \item \textbf{Average Displacement Error (ADE):}  
  \begin{equation}
    \mathrm{ADE} \;=\;\frac{1}{N}\sum_{i=1}^{N}\frac{1}{T_f}\sum_{t=1}^{T_f}\bigl\|Y_i(t) - \hat{Y}_i(t)\bigr\|_2.
  \end{equation}
  ADE measures the mean Euclidean distance between the ground‐truth trajectory \(Y_i(t)\) and the single best predicted trajectory \(\hat{Y}_i(t)\) at each future time step. By averaging over all vehicles \(i\) and time frames \(t\), ADE captures the typical per‐step deviation of the predictor and reflects its ability to track continuous motion.

  \item \textbf{Final Displacement Error (FDE):}  
  \begin{equation}
    \mathrm{FDE} \;=\;\frac{1}{N}\sum_{i=1}^{N}\bigl\|Y_i(T_f) - \hat{Y}_i(T_f)\bigr\|_2.
  \end{equation}
  FDE focuses solely on the terminal position error at the prediction horizon \(T_f\). This metric emphasizes end‐point accuracy—critical for applications such as collision avoidance or waypoint arrival—by penalizing trajectories that diverge over longer horizons.

  \item \textbf{Mean Absolute Error (MAE):}  
  \begin{equation}
    \mathrm{MAE} \;=\;\frac{1}{N}\sum_{i=1}^{N}\frac{1}{T_f}\sum_{t=1}^{T_f}\bigl\|Y_i(t) - \hat{Y}_i(t)\bigr\|_1.
  \end{equation}
  Unlike ADE, MAE uses the \(\ell_1\) norm to quantify error, which is less sensitive to outliers and provides a robust measure of average deviation. By averaging absolute differences in each coordinate, MAE complements ADE and highlights situations where large but infrequent errors might skew the Euclidean average.

  \item \textbf{Root Mean Square Error (RMSE):}  
  \begin{equation}
    \mathrm{RMSE} \;=\;\sqrt{\frac{1}{N}\sum_{i=1}^{N}\frac{1}{T_f}\sum_{t=1}^{T_f}\bigl\|Y_i(t) - \hat{Y}_i(t)\bigr\|_2^2}.
  \end{equation}
  RMSE emphasizes larger deviations by squaring the Euclidean distance before averaging, then taking the square root. .
\end{itemize}

In all formulas, \(\hat{Y}_i(t)\) denotes the predicted position selected from the single best mode—i.e.\ the hypothesis minimizing the cumulative Euclidean error over the horizon. Together, these metrics furnish a comprehensive evaluation: ADE and MAE characterize per‐step consistency, FDE quantifies long‐term drift, and RMSE highlights sensitivity to large outliers.

\subsection{Prediction Error Analysis}

To rigorously evaluate our interaction‐aware forecasting framework, we compare its trajectory‐prediction accuracy against seven state‐of‐the‐art methods—Social‐LSTM (S‐LSTM) \citep{alahi2016social}, Convolutional Social‐LSTM (CS‐LSTM) \citep{deo2018convolutional}, Planning‐informed Prediction (PiP) \citep{song2020pip}, GRIP \citep{li2019grip}, STDAN \citep{chen2022intention}, GIRAFFE \citep{wu2023graph}, RHINO \citep{wu2024hypergraph}—and a simple constant‐speed extrapolation baseline.  This suite of comparisons spans both highway (NGSIM) HighD settings, ensuring that performance gains are robust across traffic regimes.

Table~\ref{tab:prediction-errors} summarizes RMSE at prediction horizons of 10, 20, 30, 40 and 50 frames.  At short horizons (10 frames), HGT already reduces error by more than 30\% relative to the next‐best method (RHINO).  As the horizon extends to 50 frames, this gap widens: HGT attains an RMSE of 2.75 m on NGSIM and 0.41 m on HighD, compared with 2.97 m and 0.89 m for RHINO, and over 8 m and 5 m for the constant‐speed baseline.  These consistent gains underscore HGT’s ability to capture both complex vehicle dynamics and group-wise interactions: by reasoning over hyperedges, our model anticipates coordinated maneuvers that pairwise or kinematic models miss.  Moreover, the constant-speed baseline’s rapidly diverging error highlights the risk of relying on naive, linear extrapolation—a core assumption of traditional TTC.  As we will show in Section \ref{ttc}, this divergence directly translates into overly optimistic, and therefore unsafe, collision-time forecasts when using constant-velocity TTC.

\begin{table}[!t]
\footnotesize
\caption{Prediction error obtained by different models in RMSE (m)}\label{tab:prediction-errors}
\centering
\begin{tabular}{>{\centering\arraybackslash}m{1.2cm} >{\centering\arraybackslash}m{1.0cm} |>{\centering\arraybackslash}m{0.8cm} >{\centering\arraybackslash}m{0.8cm} >{\centering\arraybackslash}m{0.8cm} >
{\centering\arraybackslash}m{0.8cm} >{\centering\arraybackslash}m{1.0cm} >{\centering\arraybackslash}m{1.0cm} >{\centering\arraybackslash}m{1.1cm} |>{\centering\arraybackslash}m{1.1cm} >{\centering\arraybackslash}m{1.2cm}}
\hline
\textbf{Dataset} & \textbf{Horizon (Frame)} & S-LSTM & CS-LSTM & PiP & GRIP & STDAN & GIRAFFE & RHINO & Constant Speed & \textbf{HGT}\\
\hline
\multirow{5}{*}{\textbf{NGSIM}} & 10 & 0.65 &0.61 & 0.55 & 0.37 & 0.42 & 0.38 & 0.32 & 0.55 & \textbf{0.24}\\
 & 20 & 1.31 & 1.27 & 1.18 & 0.86 & 1.01 & 0.89 & 0.78 & 1.37 & \textbf{0.73}\\
 & 30 & 2.16 & 2.08 & 1.94 & 1.45 & 1.69 & 1.45 & \textbf{1.34} & 2.38 & 1.36\\
 & 40 & 3.25 & 3.10 & 2.88 & 2.21 & 2.56 & 2.46 & 2.17 & 3.56 & \textbf{2.08}\\
 & 50 & 4.55 & 4.37 & 4.04 & 3.16 & 3.67 & 3.24 & 2.97 & 4.90 & \textbf{2.75}\\
\hline
\multirow{5}{*}{\textbf{HighD}} & 10 & 0.22 & 0.22 & 0.17 & 0.29 & 0.19 & 0.19 & 0.19 & 0.17 & \textbf{0.04}\\
 & 20 & 0.62 & 0.61 & 0.52 & 0.68 & 0.27 & 0.42 & 0.26 & 0.35 & \textbf{0.09}\\
 & 30 & 1.27 & 1.24 & 1.05 & 1.17 & 0.48 & 0.81 & 0.42 & 0.56 & \textbf{0.16}\\
 & 40 & 2.15 & 2.10 & 1.76 & 1.88 & 0.91 & 1.13 & 0.65 & 0.81 & \textbf{0.27}\\
 & 50 & 3.41 & 3.27 & 2.63 & 2.76 & 1.66 & 1.56 & 0.89 & 1.12 & \textbf{0.41}\\
\hline
\end{tabular}
\end{table}

\begin{figure}[!t]
    \centering
    \begin{subfigure}[b]{0.48\textwidth}
        \centering
        \includegraphics[width=\textwidth]{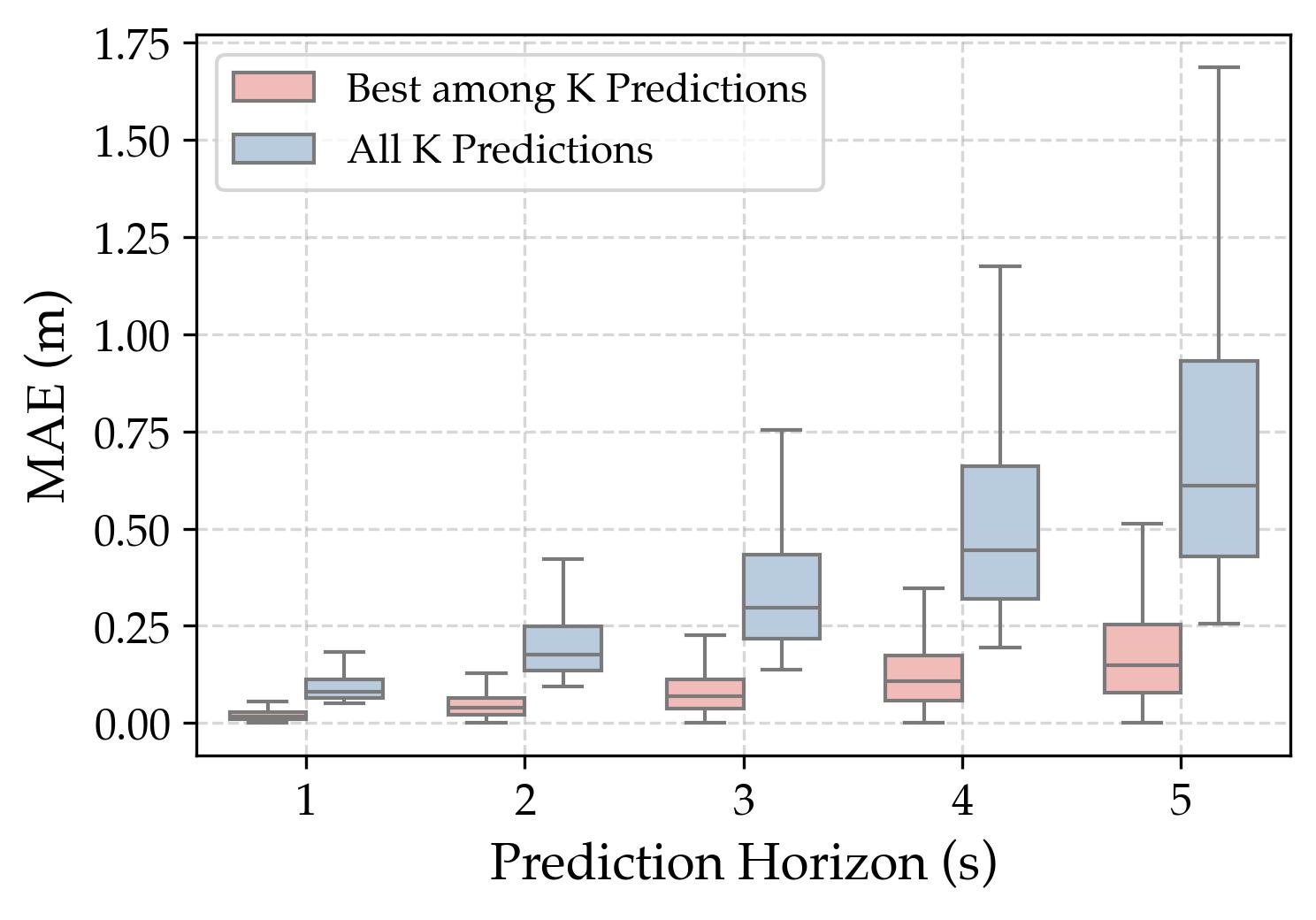}
        \caption{MAE Distribution on HighD Dataset.}
        \label{fig:prediction-error-boxplot-a}
    \end{subfigure}
    \hfill
    \begin{subfigure}[b]{0.48\textwidth}
        \centering
        \includegraphics[width=\textwidth]{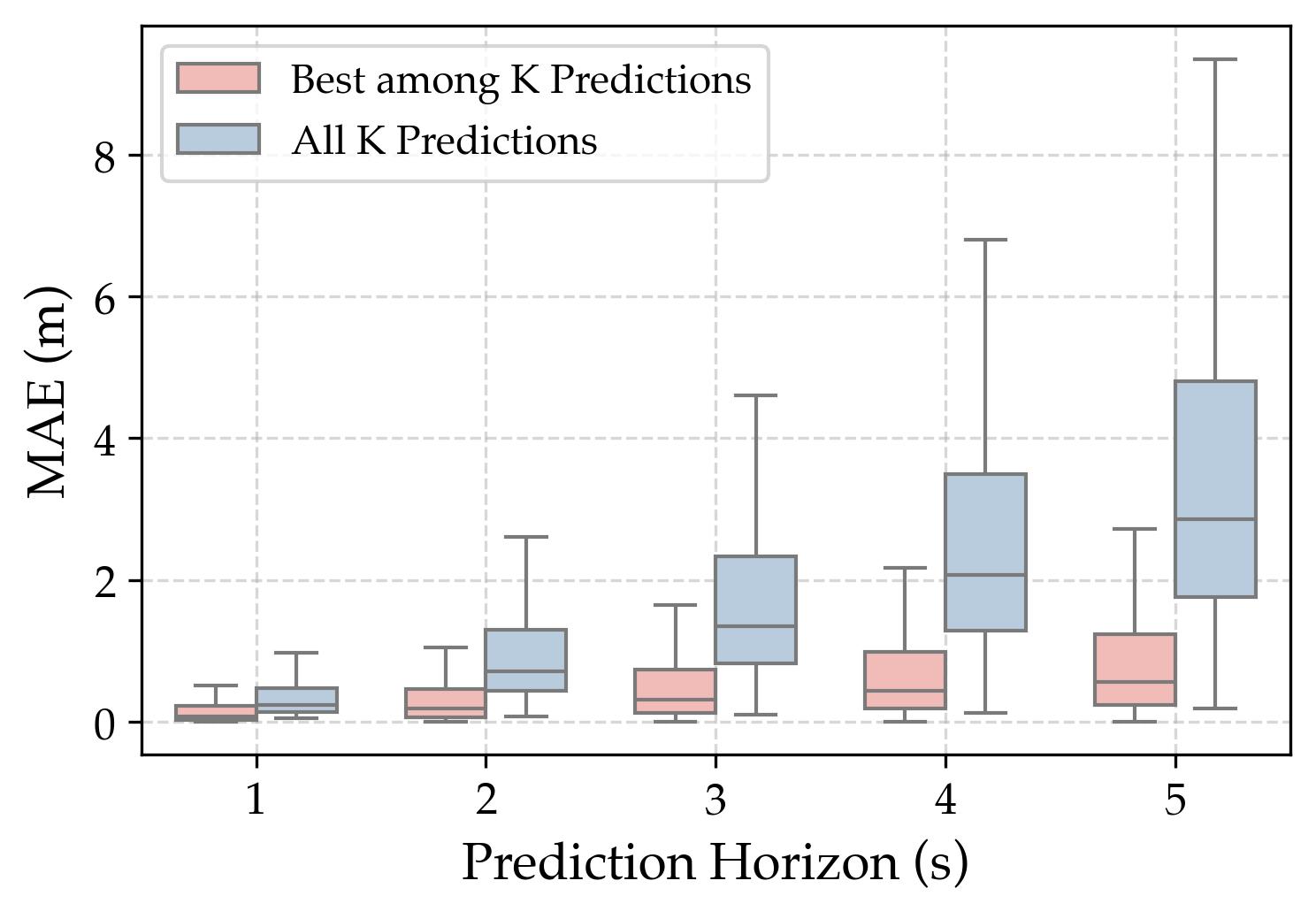}
        \caption{MAE Distribution on NGSIM Dataset.}
        \label{fig:prediction-error-boxplot-b}
    \end{subfigure}
    \caption{Comparison of positional prediction error distributions of HGT.}
    \label{fig:prediction-errors}
\end{figure}

Figure~\ref{fig:prediction-errors} further illustrates HGT’s advantages in both accuracy and uncertainty quantification.  Even when selecting the single best trajectory mode, HGT’s median MAE remains below 0.2 m on HighD and under 0.8 m on NGSIM, substantially outperforming all baselines.  The full ensemble’s tighter spread indicates that HGT captures the true variability of plausible futures without overestimating tail risk. This combination of precise point estimates and calibrated uncertainty provides a robust foundation for our stochastic HF-TTC computation, ensuring that collision‐time distributions faithfully reflect both likely and less probable—but potentially critical—trajectory deviations.

\subsection{Scenario Analysis} \label{ttc}

\begin{figure}[!ht]
    \centering
    \includegraphics[width=1.0\textwidth]{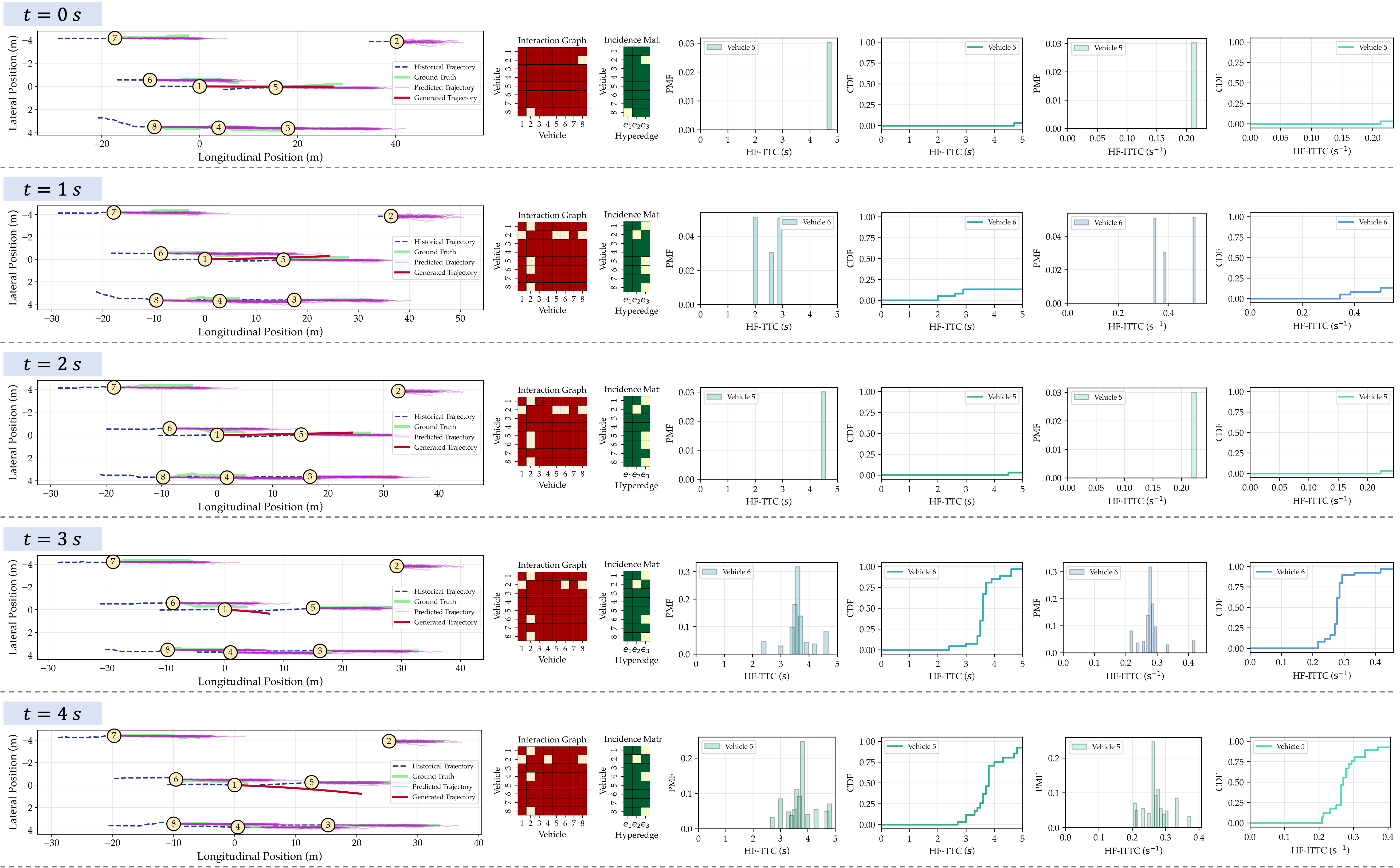}
    \caption{Scenario Analysis: host vehicle and surrounding vehicle trajectories (top row), interaction and incidence matrices (top right), PMF, CDF, time series of TTC (middle row), MF, CDF, time series of ITTC (bottom row).}
    \label{fig_exp1}
\end{figure}


To demonstrate HF-TTC’s capability for early risk detection, we analyze a prototypical highway scenario over a 4 s window (Figure~\ref{fig_exp1}), sampled at 1 s intervals. In each column, the \textit{left} panel overlays the host’s planned trajectory (red) with the ambient vehicles’ ground-truth paths (green) and probabilistic forecasts (purple); the \textit{center} panel shows the evolving hypergraph incidence matrix; and the \textit{right} panel plots the HF-TTC probability mass function (PMF, solid) and cumulative distribution function (CDF, dashed) for the two most critical neighbors, \(v_5\) and \(v_6\).

At \(t=0\) s, \(v_5\) and \(v_6\) travel at constant speed in separate lanes. Their forecast corridors are narrow, and \(H\) contains only trivial columns. Consequently, both HF-TTC PMFs peak above 4 s and their CDFs remain near zero until after 4 s—indicating ample safety buffer in free-flow conditions. By \(t=1\) s, a slight deceleration of \(v_6\) pushes its probabilistic envelope toward the host’s lane. The HF-TTC PMF for \(v_6\) shifts left to around 2 s, and its CDF accumulates mass below 3 s, signaling the first clear indication of elevated collision risk—well before the host adjusts its trajectory.  

At \(t=2\) s, the host’s RK4 planner temporarily accelerates, advancing the red trajectory forward.  Under this faster plan, \(v_5\)—which has been decelerating steadily—emerges as the dominant threat, its HF-TTC distribution now leading the risk ranking. By \(t=3\) s, sensing the tightening gap, the host switches to a more conservative behavior model, slowing and initiating a rightward turn.  Meanwhile, \(v_6\) continues to accelerate, shifting its HF-TTC PMF rightward, while \(v_5\)’s PMF remains left of it, indicating that the host’s adjustment has reprioritized the imminent conflict.

At \(t=4\) s, as \(v_6\) starts to decelerate and \(v_5\) persists in decelerating, their trajectories remain close to the host’s path. HF-TTC PMFs now concentrate between 2.5 s and 5 s, and CDFs position the bulk of probability below 4 s. The host’s slight rightward lane shift at this moment successfully re-establishes separation, validating that HF-TTC delivered timely, vehicle‐specific alerts.

These synchronized shifts in HF-TTC distributions and hypergraph connectivity reveal how our framework integrates physical motion constraints, group-wise interaction context, and probabilistic uncertainty to flag risk as soon as safety margins shrink. By separately tracking longitudinal and lateral thresholds, HF-TTC distinguishes between approaching conflicts along the road and potential cut-in maneuvers. Moreover, the hypergraph structure ensures that coordinated multi-vehicle behaviors—such as simultaneous deceleration or platoon formation—are reflected immediately in the risk metric. Together, these elements enable proactive, real-time collision warning that outpaces deterministic, pairwise TTC measures.

\subsection{Host Vehicle Behavior Model Comparison} \label{ttc}
We examine how different assumptions about the host vehicle’s future control inputs—referred to here as behavior models—affect both trajectory forecasting and collision‑risk estimation. Three data‑driven models are evaluated alongside a traditional kinematic baseline:
\begin{enumerate}[nosep]
    \item \textbf{Last-Step Constant:} The host’s most recently observed longitudinal acceleration and yaw rate are held fixed throughout the prediction horizon.
    \item \textbf{Average Constant:} The mean acceleration and yaw rate computed over the historical window are applied as constant inputs.
    \item \textbf{Self-Prediction:} Control inputs are inferred at each prediction step by replaying the host’s own predicted dynamics (i.e., continuously replanning based on the bicycle model).
    \item \textbf{Traditional TTC:} Assumes constant longitudinal and lateral velocities with no control updates.
\end{enumerate}

All models utilize the same hypergraph-based trajectory predictor and TTC computation pipeline; only the host behavior model differs. We quantify prediction accuracy using average displacement error (ADE), final displacement error (FDE), mean absolute error (MAE), and root mean squared error (RMSE), and assess safety via TTC and inverse TTC (ITTC) statistics.

\paragraph{Trajectory Prediction Accuracy} 
Table~\ref{tab:prediction_metrics} compares ADE, FDE, MAE, and RMSE for each behavior model. The Average Constant model achieves the lowest MAE (0.4216 m), while Self-Prediction closely follows with MAE = 0.4237 m. Both data‑driven models substantially outperform the Last-Step Constant model (MAE = 0.5688 m, RMSE = 0.8623 m), demonstrating the penalty incurred by ignoring control evolution. Self-Prediction yields a marginally higher RMSE (0.6555 m) than Average Constant (0.6529 m), suggesting that continuous replanning can introduce noise.

\begin{table}[ht]
    \footnotesize
    \centering
    \begin{tabular}{lcccc}
    \hline
    \textbf{Behavior Mode} & \textbf{ADE} & \textbf{FDE} & \textbf{MAE} & \textbf{RMSE} \\
    \hline
    Last Step Constant     & 0.9970 & 2.1979 & 0.5688 & 0.8623 \\
    Average Constant       & \textbf{0.7432} & \textbf{1.6861} & \textbf{0.4216} & \textbf{0.6529} \\
    Self-Prediction  & 0.7465 & 1.6918 & 0.4237 & 0.6555 \\
    Traditional TTC & 1.4850 & 3.6002 & 0.8039 & 1.3339 \\
    \hline
    \end{tabular}
    \caption{Performance comparison using ADE, FDE, MAE, and RMSE metrics in meters (m)}
    \label{tab:prediction_metrics}
\end{table}



\begin{figure}[!t]
    \centering
    \includegraphics[width=1.0\textwidth]{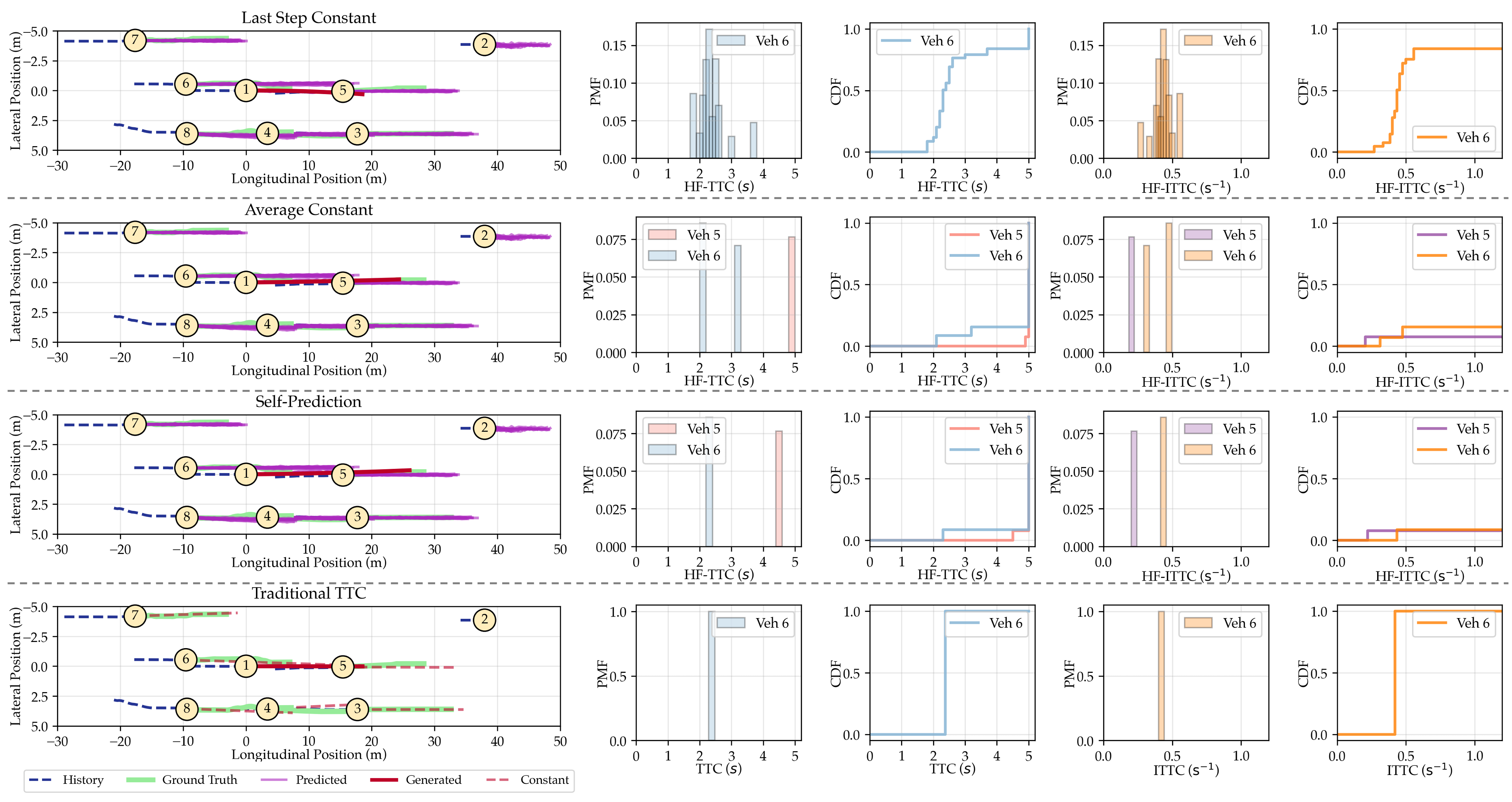}
    \caption{Comparison of host behavior models in a dense traffic scenario. \textbf{Left Column:} Trajectories for each model—Last‑Step Constant, Average Constant, Self‑Prediction, and Traditional TTC. \textbf{Right Column:} Detailed surrogate safety analyses arranged in a 3$\times$4 grid. For each model (by column), rows present: (a) TTC probability mass function (PMF), (b) TTC cumulative distribution function (CDF), (c) ITTC PMF, (d) ITTC CDF.}
    \label{fig:exp_comparison_control}
\end{figure}

\paragraph{Safety Assessment Across Behavior Models}  
Figure~\ref{fig:exp_comparison_control} contrasts host trajectories and surrogate safety metrics under Last‐Step Constant, Average Constant, Self‐Prediction, and Traditional TTC. In each left‐column panel, the host’s generated path (solid red) is shown alongside its historical trajectory (dashed blue), the ground‐truth futures (solid green), and the model’s top‐ranked predictions (semi‐transparent purple). Numbered markers identify the eight ambient vehicles; for the three data‐driven models, vehicles 5 and 6 are highlighted, while for Traditional TTC, vehicles 4 and 6 dominate the risk estimates.

Under the Last‐Step Constant model, the host vehicle decelerates and slightly drifts right compared to its prior motion, reflecting a modest control adjustment. This behavior reduces immediate longitudinal closure but still allows some lateral approach. The HF-TTC PMF for vehicle 6 peaks near 2 s, and its CDF climbs steeply between about 2 s and 3 s. The HF‐ITTC PMF concentrates between 0.25 and 0.55 s\(^{-1}\), indicating a moderate early‐warning window as the host’s small drift helps maintain separation. With the Average Constant assumption, the host vehicle largely maintains its heading and speed, proceeding almost straight through the scene. As a result, vehicle 5’s HF‐TTC PMF peaks near 5 s, while vehicle 6’s appears around 2 s, reflecting their relative lanes and speeds. The corresponding HF‐ITTC PMFs fall in the range 0.20–0.50 s\(^{-1}\)—slightly lower than in the Last‐Step model—signaling a somewhat slower accumulation of risk when the host fails to steer away. Under Self-Prediction, the host continuously replans its controls and makes a more pronounced lateral maneuver to maximize separation. In this scenario, the HF-TTC modes occur at approximately 4.5 s for vehicle 5 and 2.2 s for vehicle 6, while the HF-ITTC PMF concentrates below 0.30 s\(^{-1}\), with its CDF rising most gradually. This model achieves the largest safety buffers by adapting dynamically to emerging threats.

Unlike our stochastic HF-TTC metrics, Traditional TTC relies on a single, deterministic extrapolation—holding each vehicle’s speed and heading constant—which yields only one collision estimate per pair and offers no uncertainty quantification. In the bottom row of Figure~\ref{fig:exp_comparison_control}, Traditional TTC for vehicle 6 predicts a collision at roughly 2.3 s and its CDF jumps abruptly to one, with no gradual risk accumulation. Moreover, for other vehicles, Traditional TTC even extends predicted collision times beyond the five-second decision window, creating a false sense of safety. By contrast, our HF-TTC distributions—derived under three distinct host behavior models—span the full 2–5 s window, concentrating probability mass where risk is most critical and explicitly conveying uncertainty. This comparison illustrates how deterministic TTC systematically overestimates time to collision and underestimates imminent danger, whereas our interaction-aware, high-fidelity HF-TTC delivers earlier, probabilistic warnings that better support proactive safety interventions.

\paragraph{Distributional Insights}  

\begin{figure}[!ht]
    \centering
    \includegraphics[width=0.8\textwidth]{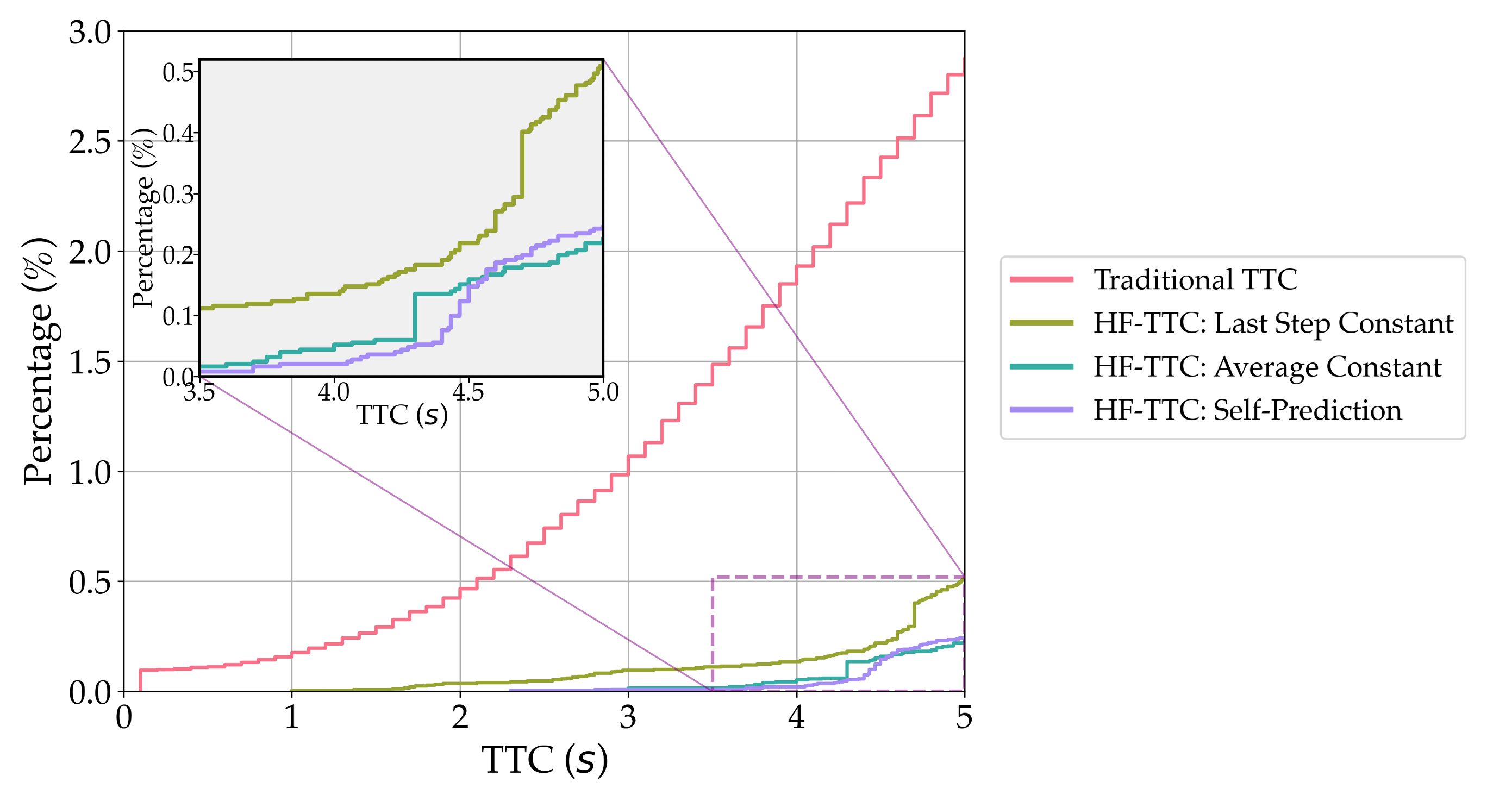}
    \caption{Cumulative distribution of TTC across the different behavior models.}    \label{fig:control_strategy_metrics}
\end{figure}

Figure~\ref{fig:control_strategy_metrics} compares the cumulative TTC distributions produced by Traditional TTC (red) and our three HF-TTC variants. Under Traditional TTC, the CDF remains below 1\% until approximately 3 s and only reaches 3\% by the 5 s horizon, suggesting a nontrivial collision probability that contradicts the well‐known rarity of highway crashes. In contrast, the HF-TTC curves rise more gradually: the Last-Step Constant model crosses just 0.1\% by 3 s and peaks around 0.5\%, while both the Average Constant and Self-Prediction variants remain below 0.24\% over the entire interval. 

These markedly lower probabilities underscore two critical points. First, deterministic, constant‐velocity TTC systematically overestimates collision likelihood and delays risk indication, potentially desensitizing safety systems and human operators to imminent threats. Second, by generating full probability distributions rather than single‐point estimates, HF-TTC faithfully represents the inherent uncertainty in multi-agent interactions and driver behavior. This probabilistic characterization enables practitioners to set informed alert thresholds, balance false alarms against missed detections, and tailor interventions to different operational contexts.

Importantly, the narrow spread and low magnitude of the HF-TTC distributions align with the empirical fact that highway collisions are rare events. By reflecting both the low base‐rate of crashes and the residual uncertainty in vehicle trajectories, our interaction-aware, high-fidelity approach supports more reliable, context‐sensitive early warnings. In doing so, it overcomes the implausibly optimistic forecasts of traditional TTC and establishes a robust foundation for proactive, risk‐based collision mitigation in complex traffic environments.

\subsection{Ablation Studies}

To rigorously quantify the individual contributions of high‑order interaction modeling, probabilistic decoding, and high‑fidelity vehicle dynamics, we designed four ablation variants and evaluated their trajectory prediction performance over multiple horizons. The variants are defined as follows:
\begin{enumerate}[nosep]
    \item \textbf{HGT-GNN:} Replaces the hypergraph transformer with a conventional Graph Neural Network (GNN), restricting interactions to pairwise edges, while preserving both probabilistic decoding and the bicycle dynamics.
    \item \textbf{HGT-Deterministic:} Retains the full hypergraph transformer and bicycle model but outputs a single deterministic trajectory per agent, eliminating uncertainty quantification.
    \item \textbf{HGT-Kinematic:} Maintains hypergraph‑based, probabilistic prediction but substitutes the high‑fidelity bicycle model with a simpler kinematic model assuming constant velocity and heading.
\end{enumerate}

We measure the root mean square error (RMSE) of predicted positions against ground truth over horizons of 10, 20, 30, 40, and 50 frames. Table~\ref{table_abalation_test} summarizes the results.

As shown in Table~\ref{table_abalation_test}, the full HGT model consistently achieves the lowest RMSE values across all prediction horizons. At the shortest horizon (10 frames), replacing the hypergraph transformer with a pairwise GNN (HGT‑GNN) increases RMSE from 0.25 m to 0.29 m, highlighting the immediate benefit of capturing higher‑order interactions. Removing uncertainty estimation (HGT‑Deterministic) yields a comparable penalty of 0.03 m, indicating that multi‑modal predictions contribute non‑negligibly even in the near term. Substituting the bicycle model with a kinematic approximation (HGT‑Kinematic) further degrades performance to 0.27 m, underscoring the importance of realistic dynamics.

As the prediction horizon extends, these gaps widen: at 50 frames, HGT‑GNN and HGT‑Deterministic incur RMSEs of 3.10 m and 3.00 m respectively, compared to 2.75 m for the full model. The HGT‑Kinematic variant exhibits an intermediate RMSE of 2.96 m. This progressive divergence confirms that high‑order interaction modeling, probabilistic decoding, and high‑fidelity dynamics each play an increasingly critical role in long‑term forecasting accuracy.

\begin{table}[!t]
\footnotesize
\caption{Ablation Test Results of \texttt{HGT} in RMSE (m)}\label{table_abalation_test}
\centering
\small
\begin{tabular}{>{\centering\arraybackslash}m{1.5cm} >{\centering\arraybackslash}m{1.5cm} >{\centering\arraybackslash}m{1.5cm} >{\centering\arraybackslash}m{1.5cm} >{\centering\arraybackslash}m{1.5cm}}
\hline
Horizon (Frame) & HGT-GNN & HGT-Deterministic & HGT-Kinematic & HGT \\
\hline
10 & 0.29 & 0.28 & 0.27 & \textbf{0.25} \\
20 & 0.95 & 0.88 & 0.83 & \textbf{0.73} \\
30 & 1.65 & 1.50 & 1.44 & \textbf{1.37} \\
40 & 2.45 & 2.25 & 2.22 & \textbf{2.08} \\
50 & 3.10 & 3.00 & 2.96 & \textbf{2.75} \\
\hline
\end{tabular}
\end{table}

Collectively, the ablation results demonstrate that the integration of group‑wise interaction awareness, uncertainty quantification, and detailed vehicle dynamics is essential for achieving state‑of‑the‑art prediction performance. By retaining all three components, the full HGT model consistently outperforms simpler alternatives, thereby laying a robust foundation for reliable collision‑risk estimation in complex, multi‑agent traffic scenarios.

\section{Conclusion}
We have presented a unified framework for surrogate safety analysis that combines high-fidelity vehicle dynamics, group-wise interaction modeling, and probabilistic trajectory forecasting to deliver rigorous, early collision warnings. Host trajectories are generated by an augmented bicycle model—incorporating steering kinematics, longitudinal acceleration, and road-grade effects—and numerically integrated with a fourth-order Runge–Kutta scheme under diverse control inputs. Simultaneously, a transformer-based hypergraph neural network learns multi-vehicle affinities to produce discrete distributions of plausible ambient trajectories. On benchmark highway datasets, our Hypergraph Transformer reduces fifty-frame RMSE to 2.75 m on NGSIM (0.22 m better than the prior best and over three times more accurate than a constant-speed baseline) and to 0.41 m on HighD (a 54 \% reduction versus the next-best model). In scenario-driven tests, our stochastic HF-TTC metric consistently issues alerts well before traditional constant-velocity TTC and constrains predicted collision probabilities below 0.5 \%, in stark contrast to TTC’s inflated estimates of up to 3 \% by the five-second mark. These results confirm that deterministic TTC not only overestimates time to collision and underrepresents imminent risk but also fails to reflect the rarity of real-world crashes, whereas HF-TTC’s interaction-aware, probabilistic outputs provide timely, credible signals that faithfully represent uncertainty.

These findings yield two key insights for surrogate safety measures. First, hyperedge-based interaction modeling uncovers collective maneuvers—such as coordinated lane changes and platoon formations—that pairwise or kinematic schemes miss, enhancing group-wise forecasting accuracy. Second, incorporating high-fidelity dynamics is essential for preserving both trajectory fidelity and safety-metric precision over longer horizons. By uniting detailed physical simulation with structured, probabilistic interaction reasoning, our framework achieves substantially earlier and more dependable collision warnings in complex traffic environments. 

Despite these advances, our current implementation makes several simplifying assumptions. Host controls are modeled as piecewise-constant inputs and hyperedge thresholds remain fixed, potentially overlooking rapid or erratic maneuvers. Environmental factors—such as variable friction, weather impacts, and sensor noise—are omitted, and the per-step Runge–Kutta integration and transformer inference introduce nontrivial computational overhead that may impede real-time deployment on resource-constrained platforms. To address these limitations, we will explore adaptive control parameterizations and dynamic thresholding for hypergraph construction, integrate environmental and perception models (e.g.\ friction, precipitation, sensor uncertainty), and optimize the pipeline for low-latency hardware. Incorporating driver-intent priors and closed-loop behavior models will further bridge simulation and real-world applications. Moreover, we plan to deploy and evaluate our framework within a high-fidelity digital twin of connected transportation infrastructure~\cite{zhang2025virtualroadssmartersafety}, using its detailed 3D environment for hardware-in-the-loop testing and calibration of our HF-TTC metrics under diverse operational conditions. This integration will enable iterative refinement of both the vehicle dynamics and interaction models, paving the way for next-generation active safety systems capable of proactive, probabilistic risk mitigation in complex, mixed-traffic environments.

\section*{\textbf{Acknowledgement:}}
This research is funded by Federal Highway Administration (FHWA) Exploratory Advanced Research 693JJ323C000010. The results do not reflect FHWA's opinions.

\bibliographystyle{ieeetr}
\bibliography{ref}

\end{document}